# Comprehensive Modeling and Question Answering of Cancer Clinical Practice Guidelines using LLMs

1st Bhumika Gupta
*Department of Electrical Engineering*
*Indian Institute of Technology Madras*
Chennai, India
bhumika.guptahc@gmail.com

2nd Pralaypati Ta
*Department of Electrical Engineering*
*Indian Institute of Technology Madras*
Chennai, India
pralaypati@htic.iitm.ac.in

3rd Keerthi Ram
*Healthcare Technology Innovation Centre*
*Indian Institute of Technology Madras*
Chennai, India
keerthi@htic.iitm.ac.in

4th Mohanasankar Sivaprakasam
*Department of Electrical Engineering*
*Indian Institute of Technology Madras*
Chennai, India
mohan@ee.iitm.ac.in

*Abstract*—The updated recommendations on diagnostic procedures and treatment pathways for a medical condition are documented as graphical flows in Clinical Practice Guidelines (CPGs). For effective use of the CPGs in helping medical professionals in the treatment decision process, it is necessary to fully capture the guideline knowledge, particularly the contexts and their relationships in the graph. While several existing works have utilized these guidelines to create rule bases for Clinical Decision Support Systems, limited work has been done toward directly capturing the full medical knowledge contained in CPGs. This work proposes an approach to create a contextually enriched, faithful digital representation of National Comprehensive Cancer Network (NCCN) Cancer CPGs in the form of graphs using automated extraction and node & relationship classification. We also implement semantic enrichment of the model by using Large Language Models (LLMs) for node classification, achieving an accuracy of 80.86% and 88.47% with zero-shot learning and few-shot learning, respectively. Additionally, we introduce a methodology for answering natural language questions with constraints to guideline text by leveraging LLMs to extract the relevant subgraph from the guideline knowledge base. By generating natural language answers based on subgraph paths and semantic information, we mitigate the risk of incorrect answers and hallucination associated with LLMs, ensuring factual accuracy in medical domain Question Answering.

## I. INTRODUCTION

Clinical Practice Guidelines (CPGs) provide clinical direction and aid the caregivers (including physicians, nurses, and pharmacists), along with patient and their families, in the decision-making process to ensure the best preventive, diagnostic, treatment, and supportive services to the patient. These are developed by multidisciplinary expert clinicians, researchers, and advocates to provide evidence-based treatment recommendations and are continuously updated with the advances in medicine. Some of the gold standard CPGs in Oncology have been developed by National Comprehensive Cancer Network (NCCN) [1], American Society of Clinical Oncology (ASCO) [2], and European Society for Medical Oncology (ESMO) [3].

These paper-based CPGs contain implicit rules that are often converted into IF-THEN statements (called *rule-base*) [4], [5] to create algorithms for making them programmatically executable by means of digital applications such as Clinical Decision Support Systems (CDSS) [6]. Several previous studies have been published in this area, aimed at embedding the rule-base into CDSS to eliminate time-consuming manual traversals and augment physicians in making quick clinical decisions [7]–[10]. However, their on-ground usage is modest because of the associated challenges, including interoperability, dependency on technology literacy, regular knowledge-base updation to match the pace with the development of medical practices and guidelines, etc. [11], [12]. One of the most crucial and fundamental challenges in CDSS lies in *the difficulty of rule-base creation and algorithm update [11] due to the involvement of cognition and human information processing* [6], [13]. In this work, we create a precise context-rich electronic rendition of paper-based CPGs, allowing medical practitioners to use this digital version of the guideline in the same way as paper-based CPGs but with improved rapid traversal. This guideline delivers the medical information contained in CPG in a timely manner, with its interpretation left at the medical practitioner's discretion. Moreover, the updates in guideline versions can easily be tracked, and the digital information database can be updated with advanced clinical practices.

Knowledge data graphs, designed to collect and propagate real-world knowledge, with nodes representing items of interest and edges denoting their relationships, have been increasingly utilized in the healthcare domain *to represent medical concepts, their relationships, to derive treatments and knowledge* from Electronic Healthcare Records, etc. Some previous works [14], [15] comprehensively represent the medical knowledge in CPGs in knowledge graphs [16] but fall short of capturing the context in its entirety. The absence of node types and meaningful relations connecting the nodes limits the understanding of the generated knowledge graph. In the proposed study, we use *the context and the content of the nodes* to classify them into one of the three most appropriate categories, viz. *disease condition*, *treatment option*, and *evaluation*, based on their properties. In addition to this, On the basis of the source and destination node labels, we assign the relations to one of the three categories, viz. *requires*, *indicates*, and *is followed by*.

Recently, Large Language Models (LLMs) have gained popularity for their advanced natural language understanding capabilities, enabling them to comprehend and interpret complex queries in a human-like manner [17]–[20]. Their effectiveness in text classification [21], [22] can be leveraged *to enhance the semantic richness of knowledge models* [23]. In this study, for automatic guideline knowledge graph augmentation, we employ *LLMs using zero-shot learning* to produce node classes exclusively based on the given prompt without any task-specific training. To further improvise the accuracy of classifications, we use *LLMs with few-shot learning* [24], [25] to generate node classifications based on the given prompt and a few annotated node classification samples.

In [15], the contents of the CPG with its extracted constraints are stored in the decision knowledge graph (DKG), forming the basis for a natural language question-answering (QA) framework that uses a deep learning (DL) model to extract the answers from DKG achieving answer generation accuracy of 67.6%. A significant shortcoming of this work is *the need for a substantial amount of high-quality training data to effectively use DL mechanisms, which can be circumvented by using LLMs* as they are pre-trained on a huge corpus.

More importantly, *the significance of accuracy in the medical domain cannot be underestimated, necessitating the demand for precise medical QA systems*. In our study, we use the power of LLMs with the integration of a reliable medical CPG knowledge model to develop a robust and accurate healthcare CPG - QA system. We overcome the significant challenge of hallucination in answer generation by LLMs, which has inhibited their adoption in the medical domain [26], by confining them only to extract information from the knowledge model [27]. This minimizes the cost and time-intensive manual labor involved in huge training dataset creation for query formulation for database searches. We generate the final answers by using the structure and semantic relationships in the subgraph extracted by querying the enriched guideline knowledge model database.

Our work captures the essence of the paper-based CPGs in *an enhanced digital replica and improves the accessibility and usability of structured medical knowledge for up-to-date evidence-based information retrieval*, simultaneously reducing the time involved in manual navigation and interpretation of paper-based CPGs.

In this research, we present a novel approach to enrich the knowledge model representation of NCCN Non-Small Cell Lung Cancer (NSCL) CPG [28] and its application to answering natural language questions using CPG knowledge base augmented LLM. The key contributions of this work are:

- We created a contextually and semantically enriched knowledge graph model of the NCCN NSCL CPG by manual labeling of the nodes and relations. While node labeling is done on the basis of the context and content of nodes, the relations are assigned labels using the rules created based on the interconnected node labels.
- We demonstrated an approach for the practical implementation of guideline knowledge model enhancement by automating the node labeling task using LLMs with zero-shot learning and few-shot learning paradigms. We compared their performance against manual annotations and report accuracy.
- We developed a factually correct evidence-based question-answering system using CPG knowledge database integrated LLMs. We used LLMs to query the semantically rich guideline knowledge base of NCCN NSCL CPG for relevant subgraph retrieval and exploited the semantics and context contained in the subgraph for answer generation.

The rest of this paper is laid out as follows. Section II discusses our methodology for CPG knowledge base enrichment and QA system development. Section III describes the technology and tools used for the implementation. We discuss the results in Section IV and conclude the study with future work prospects in Section V.

## II. METHODS

The process of the design and development of the stated Question-Answering (QA) system can be outlined in three major steps.

1) The manual enrichment of the guideline knowledge graph database.
2) The guideline model enhancement using LLMs for automated node labeling using zero-shot learning and few-shot learning.
3) The creation of NCCN cancer CPG Knowledge-Based Question-Answering system using LLMs.

These steps are elaborated in the subsequent subsections.

### A. Manual CPG knowledge graph enrichment

The cancer treatment pathways contained in the graphical flows in NCCN Non-Small Cell Lung Cancer (NSCL) CPG version 2.2024 [28] are first extracted and stored in an open standard JSON-LD format using the fully automated guideline extraction tool developed in our previous work [14]. The resultant knowledge model graph (Fig. 1), comprising nodes with guideline text interconnected by directed edges (known as relationships), is then reviewed for any false connections to ensure zero error. We removed the redundant nodes and split some nodes into further nodes to enhance the clarity of the connections in the guideline model. Additionally, we recorded the supplementary information about the node content provided by the labels positioned at the top of flows on each page of the guideline (as shown in Fig. 1) as a node property named 'context' for each of the corresponding nodes. To reinforce the context and semantics of the knowledge model even more, we labeled the nodes and relationships as detailed as follows.

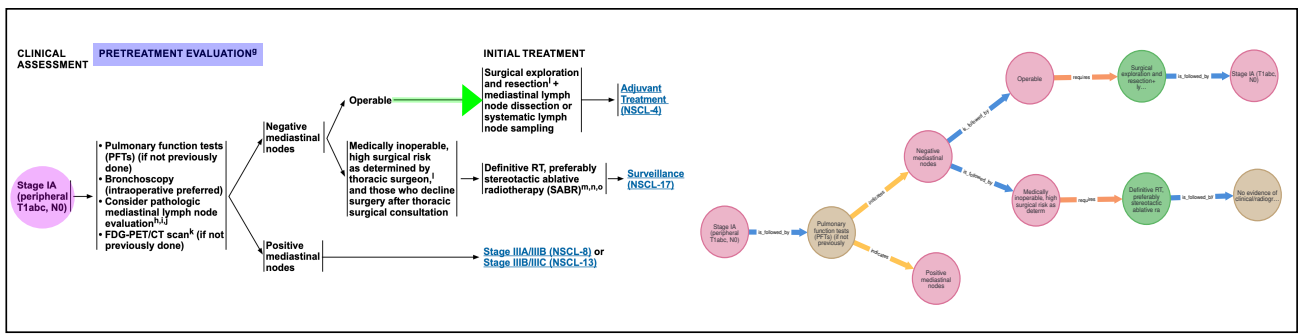

Fig. 1. A section of the enriched guideline knowledge base of NCCN NSCL CPG. The figure on the left is a guideline page highlighting the node (in the magenta-colored circle), node context (in the blue-colored rectangle), and relation (in the green-colored arrow). The figure on the right is a section from the graph knowledge base corresponding to the same guideline page imported in neo4j (a graph database management system). The pink nodes are labeled as 'Disease Condition', the green nodes as 'Treatment Option', and the light brown ones as 'Evaluation'.

*a) Node Classification:* Medical ontologies [29], [30], [31], [32], which are structured knowledge representations of collection of biomedical terminologies [33], are not particularly helpful for node classification tasks as they rely solely on text for parent class identification, with no regard for context. After careful exploration of diverse medical ontologies [34], [10], clinical articles [35], [36], healthcare information exchange standards [37] and the node texts in the guideline, we selected three classes named: '*Disease Condition*', '*Treatment Option*', and '*Evaluation*', as described below. We classified the nodes based on their text content in conjunction with the top-label context into one of these three best-suited categories.

- Treatment Option: Nodes containing text related to medical treatment procedures or medications. Examples include "Concurrent chemoradiation," "Adjuvant Systemic Therapy," and "Consider RT." etc.
- Disease Condition: Nodes providing information about disease conditions such as staging details, affected nodes information, resectability and operability status, or previous treatment data. Examples include "Progression," "Operable," "Unresectable," "N3 positive," and "Definitive local therapy possible." etc.
- Evaluation: Nodes detailing clinical evaluation procedures like scans, tests, tumor evaluations, etc. Examples include "Multidisciplinary evaluation", "PFTs, FDG-PET/CT scan, Brain MRI with contrast", "Biomarker testing", "Tumor response evaluation", etc.

Figure 1 shows the node labels for a section of the guideline.

*b) Relationship Classification:* Based on the node categories listed above and the study of medical literature, we classified the potential relationships among these node types into three distinct classes: '*requires*', '*indicates*', and '*is followed by*'. These are illustrated in Fig. 2.

The specifications of the guideline knowledge model for NCCN NSCL version 2.2024 thus obtained are described in Table I.

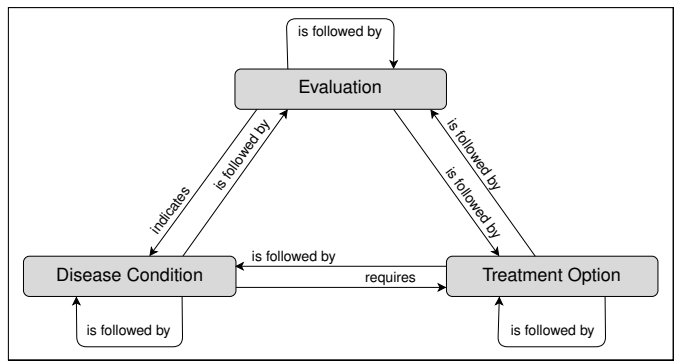

Fig. 2. Classification of relations between node types.

TABLE I
GUIDELINE KNOWLEDGE MODEL SPECIFICATIONS FOR NCCN NSCL VERSION 2.2024

| Node Type | Count | Relationship Type | Count |
|---|---|---|---|
| Disease Condition | 310 | Is followed by | 421 |
| Treatment Option | 198 | Indicates | 100 |
| Evaluation | 30 | Requires | 186 |
| Total | 538 | Total | 707 |

### B. LLM-powered Knowledge Model Enrichment

The LLM is prompted with an instruction to categorize the nodes into one of the three predetermined node classes, as mentioned in the previous subsection. Thereafter, for each node in the guideline knowledge model, the node properties (the node content and its context) are provided as inputs to the LLM to get the node label. This complete process of node classification without giving training examples (i.e., zero-shot learning) is depicted in Fig. 3. The obtained classifications are then compared and validated against the manual annotations.

Twenty-three distinct nodes from the incorrect node classifications obtained using the previous technique are given as training instances with their correct manually annotated labels to boost classification accuracy using the few-shot learning paradigm. The prompt given to the LLM is also slightly modified to include these examples. This technique

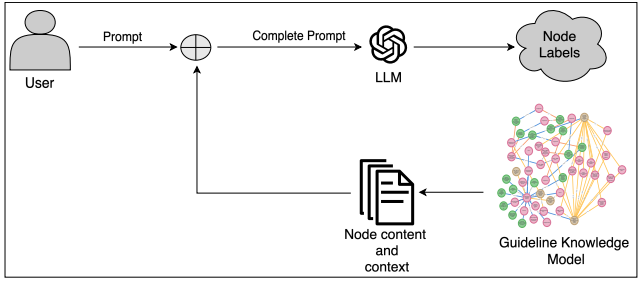

Fig. 3. Node Labeling using LLM with zero-shot learning.

is demonstrated in Fig. 4. We compared the effectiveness of both of these techniques.

## C. Knowledge Model Question Answering

We dropped the node properties that were unimportant for the purpose of question-answering (such as footnote references, t-score, m-score, prev, etc.) and imported the modified JSON-LD file containing the context-enriched guideline knowledge model in a graph database management system.

We created a dataset containing 72 natural language questions confined by guideline texts, each accompanied by its corresponding Cypher code (i.e., a structured query language used for graph database querying) [38], [39]. The dataset is broadly divided into two sets based on two different kinds of questions, as mentioned below.

*a) Set A:* It consists of 26 questions about the general treatment approach centered entirely on staging, with no particular defined disease condition. These paths can be traced from the first page of the guideline, eliminating the need for an extensive search.

*b) Set B:* It includes 46 questions concerning more specified disease conditions in addition to the cancer stage, which helps to streamline the treatment path search by searching across various node texts to match the states indicated in the question.

We further separated each of these sets into training and testing. The train set had 3 questions from set A and 10 questions from set B. The rest of the questions are added to the test set. The steps involved in the Question Answering framework are listed below.

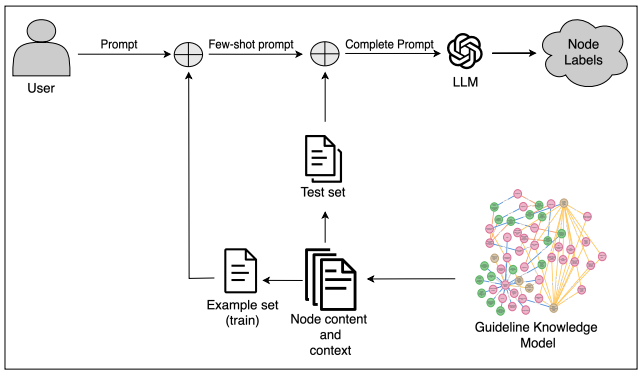

Fig. 4. Node Labeling using LLM with few-shot learning.

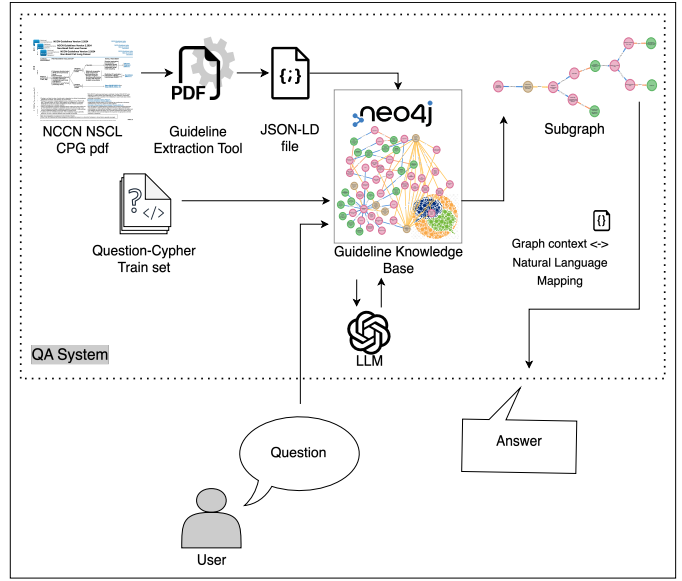

Fig. 5. NCCN NSCL CPG Question-Answering system using LLM.

1) *Providing query examples:* Query generation examples from the training question set (both from Set A and Set B), along with the schema of the guideline knowledge model (including relationship types, node types, and node properties), are given to the LLM.
2) *Providing user question:* The guideline text-constrained natural language questions from the testing set are supplied to the LLM for auto-generation of Cypher queries.
3) *Subgraph extraction:* The produced Cypher query is executed on the guideline knowledge graph database to retrieve the relevant subgraph.
4) *Answer generation:* The node contents and the semantics contained in the treatment paths obtained from the subgraph are utilized to generate factually correct, restrained vocabulary natural language answers using a graph context to natural language mapping as shown in table II.

The whole Question-Answering system developed for NCCN NSCL CPG using LLM is shown in Fig. 5.

## III. SYSTEM IMPLEMENTATION

### A. LLM-powered knowledge model enrichment

We employed the OpenAI GPT 3.5 turbo model [40] (engine gpt-3.5-turbo-instruct) via an Application Programming Interface (API) to classify the nodes using zero-shot and few-shot learning. The prompt given to the LLM for each task is as follows:

*a) Zero-Shot Learning:* To label a node of the guideline knowledge model, ⟨node text⟩ and ⟨node context⟩ in the following prompt is replaced by its content and its context (if available, otherwise, 'Not Available. Use only node text.') respectively.

TABLE II
GRAPH SEMANTICS TO NATURAL LANGUAGE ANSWER MAPPING.

| Source Node Type | Relation Type | Destination Node Type | Natural Language Text | |
|---|---|---|---|---|
| | | | For the first relation in the subgraph | For the subsequent relations in the subgraph |
| Disease Condition | requires | Treatment Option | If the disease condition is <source node text>, use the treatment <destination node text>. | If that disease condition has occurred, use the treatment <destination node text>. |
| Evaluation | is followed by | Treatment Option | Evaluate the patient for <source node text>, then use the treatment <destination node text>. | After the evaluation, use the treatment <destination node text>. |
| Treatment Option | is followed by | Treatment Option | After the treatment <source node text>is over, further use the treatment <destination node text>. | After the previous treatment is over, further use the treatment <destination node text>. |
| Disease Condition | is followed by | Evaluation | If the disease condition is <source node text>, then evaluate the patient for <destination node text>. | If that disease condition has occurred, then evaluate the patient for <destination node text>. |
| Treatment Option | is followed by | Evaluation | After the treatment <source node text>, evaluate the patient for <destination node text>. | After the previous treatment, evaluate the patient for <destination node text>. |
| Treatment Option | is followed by | Disease Condition | Check if after the treatment <source node text>, the disease condition <destination node text>has occurred. | Check if after the previous treatment, the disease condition <destination node text>has occurred. |
| Evaluation | indicates | Disease Condition | Evaluate the patient for <source node text>, check if indicates the disease condition <destination node text>. | Based on the evaluation, check if it indicates the disease condition <destination node text>. |
| Disease Condition | is followed by | Disease Condition | If the current disease condition is <source node text>, further check if the disease condition is <destination node text>. | If that disease condition has occurred, further check if the disease condition is <destination node text>. |

```
You are an expert oncologist, and you
are interpreting an NCCN Non-small cell
lung cancer guideline.
You have decided to categorise the
content in each node of the NCCN CPG
graph as either: Disease Condition,
Treatment Option, or Evaluation.
Given the following node text from
the guideline please assign the most
appropriate label among the ones
mentioned. You may use the context
whenever there is a discrepancy between
two labels but give major importance to
the node text.
node text: ⟨node text⟩
context: ⟨node context⟩
```

*b) Few-Shot Learning:* In the above zero-shot learning prompt, 'Here are some examples: ⟨examples⟩' was added before providing the node text and context of the node to be labeled for incorporating train examples for node labeling.

*B. Knowledge Model Question Answering*

Neo4j (version 4.4.5) [41], a graph database management system, was used to store the guideline knowledge model. The Open AI GPT 3.5 model [40] from the Natural Language Queries extension [42] in NeoDash [43] plugin (version 2.4.4), supported by neo4j was used to convert guideline text-constrained natural language questions into corresponding Cypher queries that reflect their intent accurately.

## IV. RESULTS AND DISCUSSION

*A. LLM-powered Node Classification*

The node classification by using LLM with zero-shot learning resulted in a classification accuracy of 80.86%. Upon using the few-shot learning of LLM, the accuracy was boosted to 88.47%. This shows that LLMs can perform the given task better when supplied with a few examples [44]. The results show that LLMs can be used to semantically strengthen the guideline knowledge model efficiently. This demonstrates the scalability of the presented guideline knowledge model enrichment approach.

*B. Knowledge Model Question Answering*

Due to hallucinations and the manner in which the natural language question is posed, LLMs, sometimes, cannot translate the natural language question into an executable Cypher query. This can be mitigated to some extent by altering the way of framing the question. In addition to that, there are multiple ways to formulate the Cypher query to retrieve the same information. We observed that the model tends to follow the provided example queries. We also encountered some minor errors in the auto-generated Cypher queries by LLM. We classified these errors in the following three types, as described below. The occurrences of which are summarized in Table III.

- Type-I: Query syntax error - Error in the Cypher query syntax generated by the LLM. For Example, the placement of a comma instead of a semicolon, the generation of Cypher codes against the schema, etc.
- Type-II: Content matching error - Mismatch in the content being searched in the auto-generated Cypher and the content present in the guideline. Improper bifurcation/merging of the guideline content in the Cypher query. For example, matching the node content 'resectable superior sulcus' in one node while the guideline contains these nodes as separate nodes, 'resectable', 'superior sulcus'.
- Type-III: Connection length error - Error in the path length to be considered to result in a subgraph. For example, increasing the search length from 4 to 7 in order to retrieve the relevant subgraph.

TABLE III
ERRORS IN AUTO-GENERATED CYPHER QUERY

| Error Type | #Occurrences in Query set A | #Occurrences in Query set B | Overall Error (%) |
|---|---|---|---|
| Type-I | 1 (4.34%) | 3 (8.33%) | 6.77 |
| Type-II | 3 (13.04%) | 8 (22.22%) | 18.64 |
| Type-III | 2 (8.69%) | 3 (8.33%) | 8.47 |
| No Error | 17 (73.91%) | 22 (61.11%) | 66.10 |
| Total | 23 | 36 | - |

Rectification of these errors requires some expertise in Cypher query formulation, information about the guideline content and familiarity with the schema of the guideline knowledge model, which serves as the shortcoming of this method.

The generated answers for one question from each of the two question sets are given as the samples below.

**Question (from Set A):** What is the treatment pathway for Stage I, central (T1abc-T2a, N0)?
**Cypher:** match (n:Disease_Condition)
where
tolower(n.content) contains "stage i"
and tolower(n.content) contains "central"
and tolower(n.context) contains "clinical stage"
with n
match path=(n)-[*2..5]->(t:Treatment_Option)
return path,nodes(path);
**Auto-Generated Cypher:**
MATCH (n:Disease_Condition)
WHERE
toLower(n.content) CONTAINS "stage i, central (t1abc-t2a, n0)" AND
toLower(n.context) CONTAINS "clinical stage"
WITH n
MATCH path=(n)-[*2..5]->(t:Treatment_Option)
RETURN path, nodes(path);
**Natural Language Answer from a path of obtained subgraph:** If the current disease condition is "Stage IB, peripheral (T2a, N0); Stage I, central (T1abc-T2a, N0); Stage II (T1abc-T2ab, N1; T2b, N0); Stage IIB (T3, N0); Stage IIIA (T3, N1)" further check if the disease condition is "Stage IB (peripheral T2a, N0) Stage I (central T1abc-T2a, N0) Stage II (T1abc-2ab, N1; T2b, N0) Stage IIB (T3, N0) Stage IIIA (T3, N1)". If that disease condition has occurred, then evaluate the patient for "Evaluate for perioperative therapy , PFTs (if not previously done) , Bronchoscopy, Pathologic mediastinal lymph node evaluation , FDG-PET/CT scan (if not previously done) , Brain MRI with contrast (Stage II, IIIA) (Stage IB [optional])". Based on the evaluation, check if it indicates the disease condition "No nodal disease". If that disease condition has occurred, further check if the disease condition is "Operable". If that disease condition has occurred, use the treatment "Surgical exploration and resection + mediastinal lymph node dissection or systematic lymph node sampling after preoperative systemic therapy, if planned".

**Question (from Set B):** What is the recommended treatment option for a Stage IIIB (T4, N2) patient with contralateral mediastinal node negative and ipsilateral mediastinal node negative?
**Cypher:** match path=(n:Disease_Condition)
-[*1..6]->(c1:Disease_Condition)
-[*1..4]->(c2:Disease_Condition)
-[*1..4]->(t:Treatment_Option)
where tolower(n.content) contains tolower("Stage IIIB (T4, N2)")
and tolower(c1.content) contains tolower('Contralateral mediastinal node negative')
and tolower(c2.content) contains tolower('ipsilateral mediastinal node negative')
return path,nodes(path),t.content
**Auto-Generated Cypher:**
MATCH path=(n:Disease_Condition)
-[*1..4]->(c1:Disease_Condition)
-[*1..4]->(c2:Disease_Condition)
-[*1..3]->(t:Treatment_Option) WHERE
toLower(n.content) CONTAINS "stage iiib (t4, n2)"
AND toLower(c1.content) CONTAINS "contralateral mediastinal node negative"
AND toLower(c2.content) CONTAINS "ipsilateral mediastinal node negative"
RETURN path, nodes(path), t.content;
**Natural Language Answer from a path of obtained subgraph:** If the disease condition is "Stage IIIB (T4, N2) Stage IIIC (T4, N3)" then evaluate the patient for "FDG-PET/CT scan (if not previously done) , Brain MRI with contrast , Pathologic confirmation of N2-3 disease by either: Mediastinoscopy Supraclavicular lymph node biopsy Thoracoscopy Needle biopsy Mediastinotomy EUS biopsy EBUS biopsy". Based on the evaluation, check if it indicates the disease condition "Contralateral mediastinal node negative". If that disease condition has occurred, further check if the disease condition is "Ipsilateral mediastinal node negative (T4, N0-1)". If that disease condition has occurred, further check if the disease condition is "Stage IIIA (T4, N0-1) unresectable". If that disease condition has occurred, use the treatment "Definitive concurrent chemoradiation (category 1)".

Table III results and examples of auto-generated Cypher queries (above) show that the LLM generates Cypher codes more precisely for generic treatment questions (73.9% accuracy in question set A) than for questions with multiple disease conditions requiring multi-hop and complex database searches (61.11% accuracy in question set B). This is naturally expected as the LLM is unaware of the guideline node contents, hence it generates Cypher queries by bifurcating the question text into chunks which do not match the guideline node content exactly. This also explains the reason why Type-II error for query set B is greater than that of query set A.

Our proposed method of question-answering on NCCN CPGs by using pre-trained LLMs requires considerably less training data (i.e., 13 instances) to generate natural language answers as opposed to 5810 instances (70% of 8300) in [15] which uses deep learning (DL) models.

To our knowledge, this work is the first attempt toward faithful knowledge representation of CPG with enriched context & semantics and its usage in accurate guideline text-restricted natural language question-answering system development.

## V. CONCLUSION

This paper presents a novel work in the area of knowledge representation and question-answering of cancer CPGs. The automatically extracted guideline knowledge model is contextually and semantically enriched, and its practical implementation is investigated using LLMs which gives promising results, as mentioned in the previous section. The presented work also provides a methodology for natural language question answering with guideline-restrained text leveraging LLMs to generate accurate answers in accordance with the guidelines. These contributions radically enhance the ease of use and comprehension of cancer guidelines, facilitating more effective utilization of these critical resources in clinical practice and decision-making.

In the future, we plan to expand this work to include over 60 additional types of cancer Clinical Practice Guidelines (CPGs) published by the NCCN. These guidelines include not only graphical care flows but also valuable recommendations presented in tables and evidence blocks. We aim to integrate this information into our existing knowledge base to ensure its completeness. Furthermore, the limitation of the current work of guideline text-restrictive question answering needs to be addressed. This will allow for a more comprehensive and nuanced understanding of user queries related to cancer treatment guidelines. Additionally, we plan to evaluate our work by comparing its performance with other open-source LLMs. This comparative analysis will provide valuable insights into the effectiveness and efficiency of our approach in comparison to existing alternatives.